# Fast and accurate sentiment classification using an enhanced Naive Bayes model.

Vivek Narayanan[1], Ishan Arora[2], Arjun Bhatia[3]

Department of Electronics Engineering,
Indian Institute of Technology (BHU), Varanasi, India

[1]`vivek.narayanan.ece09@iitbhu.ac.in`
[2]`ishan.arora.ece09@iitbhu.ac.in`
[3]`arjun.bhatia.ece09@iitbhu.ac.in`

**Abstract.** We have explored different methods of improving the accuracy of a Naive Bayes classifier for sentiment analysis. We observed that a combination of methods like effective negation handling, word n-grams and feature selection by mutual information results in a significant improvement in accuracy. This implies that a highly accurate and fast sentiment classifier can be built using a simple Naive Bayes model that has linear training and testing time complexities. We achieved an accuracy of 88.80% on the popular IMDB movie reviews dataset. The proposed method can be generalized to a number of text categorization problems for improving speed and accuracy.

## 1 Introduction

Among the most researched topics of natural language processing is sentiment analysis. Sentiment analysis involves extraction of subjective information from documents like online reviews to determine the polarity with respect to certain objects. It is useful for identifying trends of public opinion in the social media, for the purpose of marketing and consumer research. It has its uses in getting customer feedback about new product launches, political campaigns and even in financial markets [14]. It aims to determine the attitude of a speaker or a writer with respect to some topic or simply the contextual polarity of a document. Early work in this area was done by Turney and Pang ([2], [7]) who applied different methods for detecting the polarity of product and movie reviews.

Sentiment analysis is a complicated problem but experiments have been done using Naive Bayes, maximum entropy classifiers and support vector machines. Pang et al. found the SVM to be the most accurate classifier in [2]. In this paper we present a supervised sentiment classification model based on the Naïve Bayes algorithm.

Naïve Bayes is a very simple probabilistic model that tends to work well on text classifications and usually takes orders of magnitude less time to train when compared to models like support vector machines. We will show in this paper that a high degree of accuracy can be obtained using Naïve Bayes model, which is comparable to the current state of the art models in sentiment classification.

## 2     Data

We used a publicly available dataset of movie reviews from the Internet Movie Database (IMDb) [1] which was compiled by Andrew Maas et al. It is a set of 25,000 highly polar movie reviews for training, and 25,000 for testing. Both the training and test sets have an equal number of positive and negative reviews. We chose movie reviews as our data set because it covers a wide range of human emotions and captures most of the adjectives relevant to sentiment classification. Also, most existing research on sentiment classification uses movie review data for benchmarking.

We used the 25,000 documents in the training set to build our supervised learning model. The other 25,000 were used for evaluating the accuracy of our classifier.

## 3     Naïve Bayes Classifier

A Naive bayes classifier is a simple probabilistic model based on the Bayes rule along with a strong independence assumption.

The Naïve Bayes model involves a simplifying conditional independence assumption. That is given a class (positive or negative), the words are conditionally independent of each other. This assumption does not affect the accuracy in text classification by much but makes really fast classification algorithms applicable for the problem. Rennie et al discuss the performance of Naïve Bayes on text classification tasks in their 2003 paper. [6]

In our case, the maximum likelihood probability of a word belonging to a particular class is given by the expression:

$$P(x_i | c) = \frac{Count\ of\ x_i\ in\ documents\ of\ class\ c}{Total\ no\ of\ words\ in\ documents\ of\ class\ c} \quad (1)$$

The frequency counts of the words are stored in hash tables during the training phase.

According to the Bayes Rule, the probability of a particular document belonging to a class $c_i$ is given by,

$$P(c_i|d) = \frac{P(d\ |c_i) * P(c_i)}{P(d)} \quad (2)$$

If we use the simplifying conditional independence assumption, that given a class (positive or negative), the words are conditionally independent of each other. Due to this simplifying assumption the model is termed as "naïve".

$$P(c_i|d) = \frac{(\prod P(x_i|c_j)) * P(c_j)}{P(d)} \quad (3)$$

Here the $x_i$ s are the individual words of the document. The classifier outputs the class with the maximum posterior probability.

We also remove duplicate words from the document, they don't add any additional information; this type of naïve bayes algorithm is called Bernoulli Naïve Bayes. Including just the presence of a word instead of the count has been found to improve performance marginally, when there is a large number of training examples.

## 4    Laplacian Smoothing

If the classifier encounters a word that has not been seen in the training set, the probability of both the classes would become zero and there won't be anything to compare between. This problem can be solved by Laplacian smoothing

$$P(x_i|c_j) = \frac{Count(x_i) + k}{(k+1) * (No\ of\ words\ in\ class\ c_j)} \quad (4)$$

Usually, k is chosen as 1. This way, there is equal probability for the new word to be in either class. Since Bernoulli Naïve Bayes is used, the total number of words in a class is computed differently. For the purpose of this calculation, each document is reduced to a set of unique words with no duplicates.

## 5    Negation Handling

Negation handling was one of the factors that contributed significantly to the accuracy of our classifier. A major problem faced during the task of sentiment classification is that of handling negations. Since we are using each word as feature, the word "good" in the phrase "not good" will be contributing to positive sentiment rather that negative sentiment as the presence of "not" before it is not taken into account.

To solve this problem we devised a simple algorithm for handling negations using state variables and bootstrapping. We built on the idea of using an alternate representation of negated forms as shown by Das & Chen in [3]. Our algorithm uses a state variable to store the negation state. It transforms a word followed by a not or n't into "not_" + word. Whenever the negation state variable is set, the words read are treated as "not_" + word. The state variable is reset when a punctuation mark is encountered or when there is double negation. The pseudo code of the algorithm is described below:

*PSEUDO CODE:.*

```
negated := False
for each word in document:
   if negated = True:
       Transform word to "not_" + word.
   if word is "not" or "n't":
       negated := not negated
   if a punctuation mark is encountered
       negated := False.
```

Since the number of negated forms might not be adequate for correct classifications. It is possible that many words with strong sentiment occur only in their normal forms in the training set. But their negated forms would be of strong polarity.

We addressed this problem by adding negated forms to the opposite class along with normal forms of all the features during the training phase. That is to say if we encounter the word "good" in a positive document during the training phase, we increment the count of "good" in the positive class and also increment the count of "not_good" for the negative class. This is to ensure that the number of "not_" forms are sufficient for classification. This modification resulted in a significant improvement in classification accuracy (about 1%) due to bootstrapping of negated forms during training. This form of negation handling can be applied to a variety of text related applications.

## 6    n - grams

Generally, information about sentiment is conveyed by adjectives or more specifically by certain combinations of adjectives with other parts of speech.

This information can be captured by adding features like consecutive pairs of words (bigrams), or even triplets of words (trigrams). Words like "very" or "definitely" don't provide much sentiment information on their own, but phrases like "very bad" or "definitely recommended" increase the probability of a document being negatively or positively biased. By including bigrams and trigrams, we were able to capture this information about adjectives and adverbs. Using bigrams and trigrams require a substantial amount of data in the training set, but this is not a problem as our training set had 25,000 reviews. But the data may not be enough to add 4-grams, as this may over-fit the training set. The counts of the n-grams were stored in a hash table along with the counts of unigrams.

## 7    Feature Selection

Feature selection is the process of removing redundant features, while retaining those features that have high disambiguation capabilities.

The use of higher dimensional features like bigrams and trigrams presents a problem, that of the number of features increasing from 300,000 to about 11,000,000. Most of these features are redundant and noisy in nature. Including them would affect both efficiency and accuracy. A basic filtering step of removing the features/terms which occur only once is performed. Now the number of features is reduced to about 1,500,000 features. The features are then further filtered on the basis of mutual information [3].

### 7.1 Mutual Information

Mutual information is a quantity that measures the mutual dependence of the two random variables. Formally, the mutual information of two discrete random variables *X* and *Y* can be defined as:

$$I(X;Y) = \sum_{y \in Y} \sum_{x \in X} p(x,y) \log \left( \frac{p(x,y)}{p(x)\,p(y)} \right), \quad (5)$$

Where p(x,y) is the joint probability distribution function of X and Y, and P(X) and P(Y) are the marginal probability distribution functions of X and Y respectively. Here X is an individual feature which can take two values, the feature is present or absent and Y is the class, positive or negative. We selected the top k features with maximum mutual information. By plotting a graph between accuracy and number of features, the optimal value for k was found out to be 32,000.

A plot of **Accuracy versus Number of features** is shown in Fig 1:

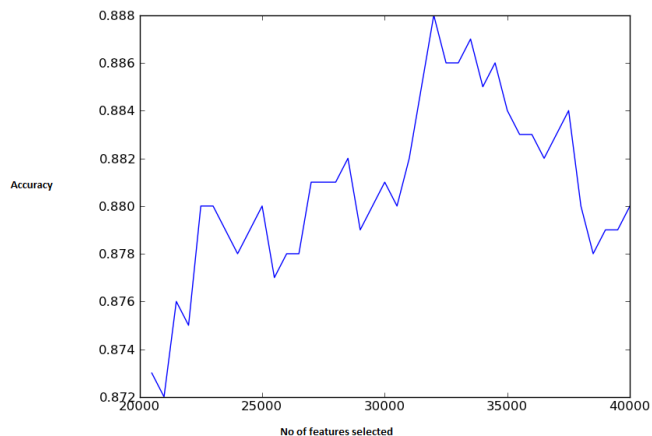

**Fig. 1.** Plot of Accuracy v/s No of features selected on Validation set

## 8 Results

We implemented the classifier in Python using hash tables to store the counts of words in their respective classes. The code is available at [13]. Training involved preprocessing data and applying negation handling before counting the words. Since we were using Bernoulli Naive Bayes, each word is counted only once per document. On a laptop running an Intel Core 2 Duo processor at 2.1 GHz, training took around 1 minute 30 seconds and used about 700 megabytes of memory. The memory usage stems largely from bigrams and trigrams prior to feature selection.

The optimal number of features was chosen by using a validation set of 1000 documents, the plot of accuracy v/s number of features is shown in Fig 1. Then the accuracy was measured on the entire test set of 25000 documents. The time for feature selection was about 3 minutes.

### 8.1 Performance and Comparison

We obtained an overall classification accuracy of 88.80% on the test set of 25000 movie reviews. The running time of our algorithm is $O(n + V \lg V)$ for training and $O(n)$ for testing, where n is the number of words in the documents (linear) and V the size of the reduced vocabulary. It is much faster than other machine learning algorithms like Maxent classification or Support Vector Machines which take a long time to converge to the optimal set of weights. The accuracy is comparable to that of the current state-of-the-art algorithms used for sentiment classification on movie reviews. It achieved a better or similar accuracy when compared to more complicated models like SVMs, autoencoders, contextual valence shifters, matrix factorisation, appraisal groups etc used in [2], [7], [8], [9], [10], [11] on the dataset of IMDb movie reviews.

### 8.2 Results Timeline

The table and graph illustrate the evolution of the accuracy of our classifier and how the inclusion of certain features helped.

**Table 1.** RESULTS TIMELINE

| Feature Added | Accuracy on test set |
|---|---|
| Original Naive Bayes algorithm with Laplacian Smoothing | 73.77% |
| Handling negations | 82.80% |
| Bernoulli Naive Bayes | 83.66% |
| Bigrams and trigrams | 85.20% |
| Feature Selection | 88.80% |

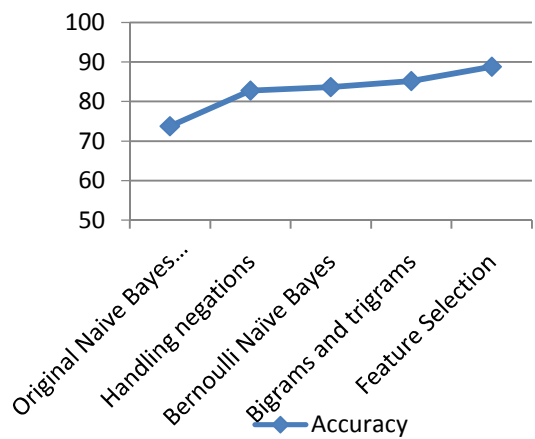

**Fig. 2.** Evolution of classification accuracy.

## 9   Conclusion

Our results show that a simple Naive Bayes classifier can be enhanced to match the classification accuracy of more complicated models for sentiment analysis by choosing the right type of features and removing noise by appropriate feature selection. Naive Bayes classifiers due to their conditional independence assumptions are extremely fast to train and can scale over large datasets. They are also robust to noise and less prone to overfitting. Ease of implementation is also a major advantage of Naive Bayes. They were thought to be less accurate than their more sophisticated counterparts like support vector machines and logistic regression but we have shown through this paper that a significantly high accuracy can be achieved. The ideas used in this paper can also be applied to the more general domain of text classification.

## Acknowledgement

We would like to express our sincere gratitude to our mentor, Professor R. R. Das for sparing his valuable time and guiding us during the project. We are also thankful to Prof. P. K. Mukherjee for his support and encouragement. Lastly, we would like to thank all the faculty members and support staff of the Department of Electronics Engineering, IIT (BHU).